\documentclass[journal]{IEEEtran}

\usepackage{ifpdf}
\usepackage{cite}
\usepackage{amsmath,amssymb,amsfonts}
\usepackage{algorithmic}
\usepackage{graphicx, adjustbox}
\usepackage{array}
\usepackage{enumitem}
\usepackage{breqn}
\usepackage{ dsfont }
\usepackage{ upgreek }
\usepackage{textcomp}
\usepackage{multirow}
\usepackage{algorithm}
\usepackage{algorithmic}
\usepackage{caption}
\usepackage{pgfplots}
\usepackage{tkz-euclide}
\usepackage{subcaption}
\usepackage{tikz}{\tiny }
\usepackage{hyperref}
\usepackage{balance}
\usepackage{bigints}

\usepackage{multirow}
\newcolumntype{P}[1]{>{\centering\arraybackslash}p{#1}}

\begin{document}
	
	\title{Federated Learning for Industrial Internet of Things \\ in Future Industries}
	
	\author{Dinh C. Nguyen,	Ming Ding, Pubudu N. Pathirana,
		Aruna Seneviratne,  Jun Li, \\ Dusit Niyato,~\IEEEmembership{Fellow,~IEEE}, and H. Vincent Poor,~\IEEEmembership{Fellow,~IEEE}
		
		\thanks{Dinh C. Nguyen and Pubudu N. Pathirana are with School of Engineering, Deakin University, Australia}
		\thanks{Ming Ding is with Data61, CSIRO, Australia}
		\thanks{Aruna Seneviratne is with School of Electrical Engineering and Telecommunications, University of New South Wales, Australia}
		\thanks{Jun Li is with School of Electrical and Optical Engineering, Nanjing University of Science and Technology, China}
		\thanks{Dusit Niyato is with School of Computer Science and Engineering, Nanyang Technological University, Singapore}
		\thanks{H. Vincent Poor is with the Department of Electrical Engineering, Princeton University, USA.}
		
	}
	
	\markboth{Accepted at IEEE Wireless Communications Magazine}%
	{}

	\maketitle
	
	\begin{abstract}
	\textcolor{black}{The Industrial Internet of Things (IIoT) offers promising opportunities to transform the operation of industrial systems and becomes a key enabler for future industries.} Recently, artificial intelligence (AI) has been widely utilized for realizing intelligent IIoT applications where AI techniques require centralized data collection and processing. However, this is not always feasible in realistic scenarios due to the high scalability of modern IIoT networks and \textcolor{black}{growing industrial data confidentiality}. \textcolor{black}{Federated Learning (FL), as an emerging collaborative AI approach,} is particularly attractive for intelligent IIoT networks by coordinating multiple IIoT devices and machines to perform AI training at the network edge while helping protect user privacy. In this article, we provide a detailed overview and discussions of the emerging applications of FL in key IIoT services and applications. A case study is also provided to demonstrate the feasibility of FL in IIoT. Finally, we highlight a range of interesting open research topics that need to be addressed for the full realization of FL-IIoT in industries.
	\end{abstract}
	
	\begin{IEEEkeywords}
		Federated learning, Industrial Internet of Things, Industry 4.0, future industries, privacy.  
	\end{IEEEkeywords}
	
	\IEEEpeerreviewmaketitle

\section{Introduction}
\textcolor{black}{Recent advances in communication and smart device technologies along with the rapid development of industrial informatization have promoted the proliferation of the Industrial Internet of Things (IIoT), with its capability to increase productivity and efficiency in industries \cite{updated1}. It is anticipated that IIoT will play an increasingly significant role in the development of new applications, from smart manufacturing, smart factory to smart transportation and smart healthcare in the future industrial revolutions, including Industry 4.0. For example, IIoT can provide innovative solutions to drive smart manufacturing processes due to its ubiquitous sensing and computation capabilities. }

To realize intelligent IIoT services and applications in industries, artificial intelligence (AI) techniques such as machine learning (ML) have been widely exploited to train data models. Traditionally, AI functions are placed at the cloud or the data center for data learning and modeling, which remains some critical limitations with respect to the rapid increase in IIoT data volumes. The transfer of a massive volume of IIoT data to a remote server for AI training requires much network bandwidth and incurs high communication overhead, both of which are unacceptable to time-sensitive IIoT applications such as autonomous driving and real-time healthcare. Importantly, the reliance on such a central server or third party for data learning raises critical privacy issues, e.g., user information leakage, since these data may contain sensitive information. Moreover, in the future industries, such a centralized AI architecture may be no longer suitable because IIoT data are not centrally located, but distributed over a large-scale network. Therefore, there is an urgent need to go toward distributed AI approaches for enabling scalable and privacy-promoting intelligent IIoT applications at the network edge.

Recently, federated learning (FL) \cite{3} has emerged as a promising solution for realizing cost-effective intelligent IIoT applications with improved privacy protection. \textcolor{black}{Conceptually, FL is a collaborative AI approach that enables training of high-quality AI models by averaging local updates aggregated from multiple learning clients, e.g., IIoT devices, without the need for direct access to the local data which thus mitigates privacy leakage risks.} Moreover, since FL attracts large computation and dataset resources from a number of IIoT devices to train AI models, the IIoT data training quality, e.g., accuracy, would be significantly improved which might not be achieved by using centralized AI approaches with less data and limited computational capabilities \cite{revise2}.
\begin{table*}
	\centering
	\caption{\textcolor{black}{Comparison with related work and new contributions of this article. }}
	\label{Table:Comparisons}
	{\color{black}
		\setlength{\tabcolsep}{5pt}
		\begin{tabular}{|P{2cm}|P{3cm}|P{10cm}|}
			\hline
			\centering \textbf{Related works}& 
			\centering \textbf{Topic}&	
			\textbf{Key contributions}
			\\
			\hline

			\cite{3}&	FL for cognitive computing in Industry 4.0&	An FL-based solution  for  big  data  driven  cognitive  computing in Industry 4.0, aiming to improve the performances of poisoning attack resistance, accuracy, and incentive mechanisms for industrial automation. 
			\\
			\hline
			\cite{revise2}&	FL for industrial artificial intelligence&	An efficient and privacy-enhanced FL solution for industrial artificial intelligence with user privacy awareness. 
			\\
			\hline
			\cite{revise3}&	FL and digital twin empowered IIoT&	An architecture of digital twin-enabled IIoT, by using digital twins to capture the characteristics of industrial devices to assist FL. 
			\\
			\hline
			This article&FL and IIoT in industries&	A holistic discussion of the use of FL in IIoT. Particularly, 
			\begin{itemize}
				\item We identify and discuss the potential of FL in various key IIoT services, i.e., IIoT data offloading and caching, IIoT attack detection, and mobile IIoT crowdsensing. 
				\item	We then provide a detailed investigation of using FL in four important applications, namely smart manufacturing, smart transportation, smart grid, and smart healthcare. 
				\item	We then present a case study toward FL-based smart healthcare. Several interesting open research topics for FL-IIoT in industries are also highlighted. 
			\end{itemize}
			\\
			\hline
	\end{tabular}}
\end{table*}

Motivated by these appealing characteristics, a flurry of research activities combining FL with IIoT in industries has been sparked \cite{3,revise2,revise3}. However, these works only focus on certain application domains in IIoT, such as cognitive computing \cite{3}, industrial artificial intelligence \cite{revise2}, and digital twin-enabled IIoT \cite{revise3},  while \textcolor{black}{a holistic overview} on the use of FL in key IIoT services and applications is still missing. 

\textcolor{black}{To fill this gap, this article presents and details an integration of FL and IIoT. Specifically, we present the principle of FL and explain its benefits in IIoT. Then, we provide the state-of-the-art review of the use of FL in important IIoT services,  i.e., IIoT data offloading and caching, IIoT attack detection, and IIoT mobile  crowdsensing. Notably, we explore and discuss the roles of FL in key industrial IIoT applications, including smart manufacturing, smart transportation, smart grid, and smart healthcare. A case study is provided in the area of federated healthcare to demonstrate the feasibility of FL in IIoT. We also highlight interesting open issues for future FL-IIoT research. } The new contributions of this article compared to the state-of-the-art are summarized in Table~\ref{Table:Comparisons}.

\textcolor{black}{The remainder of the article is organized as follows. Section~\ref{Section:RecentAdvance} describes the key principle of FL and its benefits in IIoT. We then present the overview of the potential of FL in IIoT services in Section~\ref{Section:Service}. In Section~\ref{Section:App}, we provide an detailed discussion on the integration of FL into key IIoT applications. A case study in FL-based smart healthcare is provided in Section~\ref{Section:Case}. Finally, Section~\ref{Section:Conclude} concludes the article.}

\section{Integration of FL and IIoT: Key Principle and Benefits}
\label{Section:RecentAdvance}
\subsection{Key Principle}
\begin{figure*}
	\centering
	\includegraphics[width=0.9\linewidth]{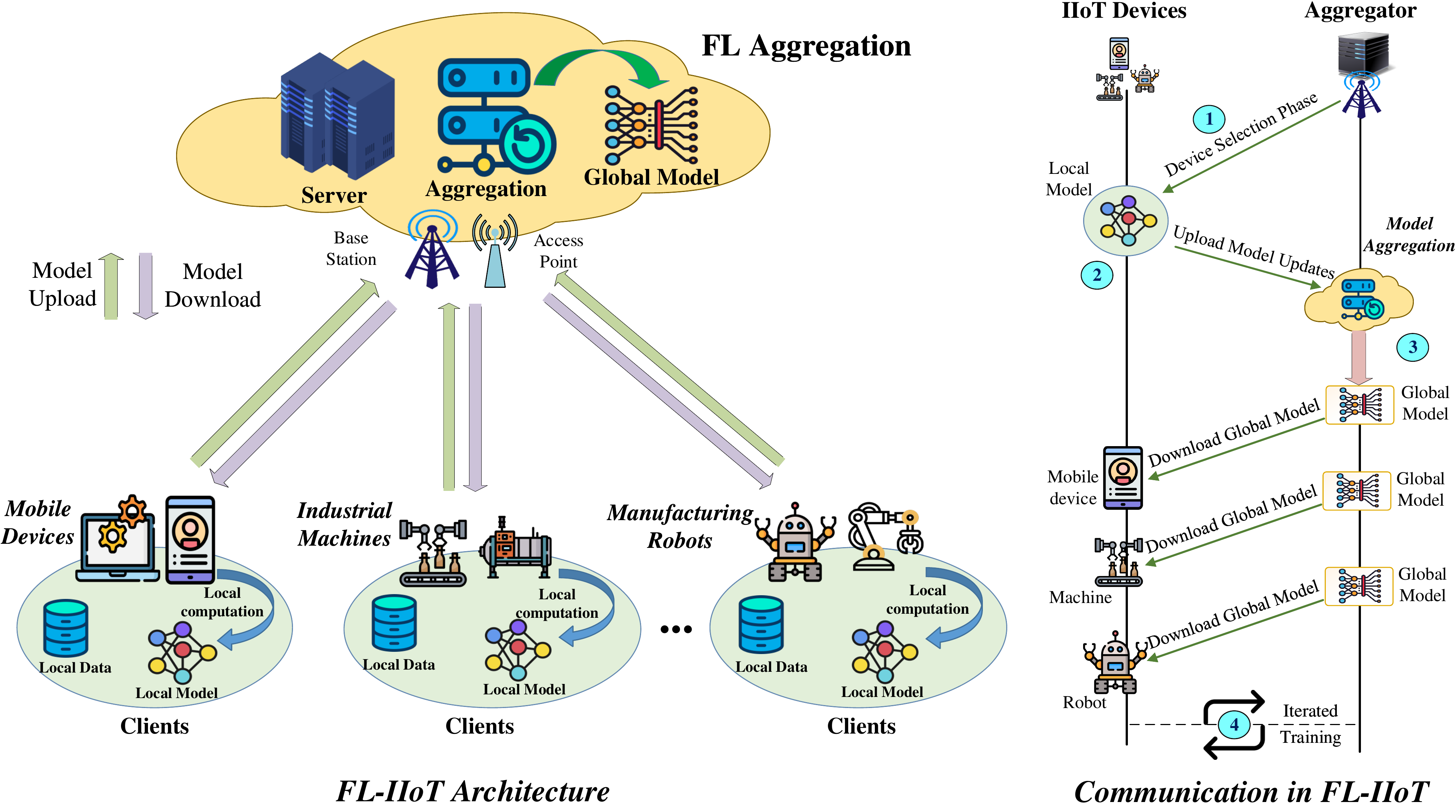}
	\caption{\textcolor{black}{The network architecture and communication process for FL-IIoT. }}
	\label{Fig:FL_Concept}
	\vspace{-0.1in}
\end{figure*}

\textcolor{black}{The typical FL-IIoT network is composed of two main entities: the data clients, e.g., IIoT devices and industrial sensors, and an aggregator \textcolor{black}{(e.g., an edge server)} located at a base station (BS) or an access point (AP), as illustrated in Fig.~\ref{Fig:FL_Concept}. FL allows IIoT devices and the server to train a shared global model while the raw data are kept at local devices. Here, each IIoT user participates in training a shared AI model by using its own dataset and then uploads its local model to the aggregator for building a new global model. By relying on the distributed data training at IIoT devices, the aggregation server can enrich the training performance without completely compromising user data privacy \cite{revise2}.} As shown in Fig.~\ref{Fig:FL_Concept}, the generic FL-IIoT process includes the following key steps:
\begin{enumerate}
	\item 	\textit{System Initialization and Device Selection:} The aggregation server selects an IIoT task, e.g., road traffic evaluation or healthcare analytic, along with model requirements such as task classification or task prediction, and learning parameters such as learning rate. Moreover, the server selects a subset of IIoT devices as the learning clients that should be involved in the FL process.
	\item	\textit{Distributed Local Training and Updates:} Once the subset of the learning clients is determined, the server sends an initial model to the clients to trigger the distributed training. In every communication round, each client trains a local model using its own dataset and calculates an update. Then, each client uploads its computed update to the server for aggregation.
	\item	\textit{Model Aggregation and Download:} After receiving all updates from clients, the server aggregates them and calculates a new version of global model. Subsequently, the server broadcasts the new global update to all clients for optimizing the local models in the next learning round. 
	\item \textit{Iterated Training:} The FL training is iterated until the global loss function converges or a desired accuracy is achieved. \textcolor{black}{Here, the accuracy of FL can be defined as the ratio of total accuracies of all clients to the total number of clients, according to the popular Federated Averaging (FedAvg) algorithm proposed by Google \cite{revise2}.} 
\end{enumerate}
\subsection{Key Benefits of FL Integration in IIoT}
With its innovative operational concept, FL can offer some important benefits for IIoT applications in industries as follows:
\begin{itemize}
	\item \textit{Data Privacy Enhancement:} In the FL system, only the local updates are required by the central server for the AI training, while the local data are kept at local devices, which thus provides a degree of data privacy. Following the increasingly stringent data privacy protection legislation such as the General Data Protection Regulation (GDPR), the capability of protecting user information of FL is significant for building sustainable and safe IIoT systems.
	\item	\textit{Low-latency Network Communication:} By avoiding the offloading of huge data volumes to the server, FL can significantly reduce communication costs in intelligent IIoT networks, e.g., latency, consumed by raw data transmission. Therefore, FL also helps save much network spectrum resources required for data training.
	\item	\textit{Improved Learning Quality:} FL attracts large computation and dataset resources from a number of IIoT devices over the distributed IIoT network to train AI models. This cooperation would accelerate the convergence rate of the overall training process and improve learning accuracy, which might not be achieved by using centralized AI approaches. 
\end{itemize}
\textcolor{black}{Compared to traditional distributed learning \cite{reviewer21, reviewer22}, which mostly performs parallel data training without federation, FL can better exploit  similar experienced data from distributed data sources located at distributed IIoT devices, which might otherwise result in ignoring rarely occurring yet important exemplars. Hence, FL is able to gain benefits from data feature diversity across the distributed dataset which helps improve the generalizability of the global AI model for better training performance, e.g., enhanced training accuracy. }


\section{FL for IIoT Services}
\label{Section:Service}
\subsection{FL for the Optimization of IIoT Data Offloading and Caching}
\textcolor{black}{To meet the ever-increasing computation demands of IIoT users and industrial operators in Industry 4.0, data offloading has been widely regarded as an efficient solution which enables IIoT devices and machines to offload their data tasks to resourceful edge servers. In this context, FL can be used to implement offloading optimization where multiple IIoT devices like actuators in smart manufacturing work as intelligent agents to collaboratively train an AI model to learn the policy of offloading industrial data, e.g., production-related data packages. This solution not only enhances data privacy due to the distribution of data learning in different IIoT devices but also mitigates the computation burden posed on the industrial system in the centralized offloading architecture.} For example, FL can support data offloading for the transportation industry \cite{6}, where each vehicle collaboratively performs data training for offloading optimization. It prevents sharing actual data and thus helps address privacy concerns of vehicle drivers. 

Data offloaded from IIoT devices can be cached by edge servers where FL can play an important role in establishing intelligent caching policies, in order to cope with the explosive growth of industrial data in modern IIoT networks. As shown in \cite{revise1}, FL is very useful to build proactive data caching schemes in urban informatics where an IIoT-based transportation system is created by the federation of vehicular entities, including macrobase stations, road side units, and moving vehicles. Here, each vehicle equipped with caching resources trains a local model using a noise-added gradient-descent algorithm and collaborates with other entities to build a shared content caching policy. 

\subsection{FL for IIoT Attack Detection}

\begin{figure*}
	\centering
	\includegraphics[width=0.9\linewidth]{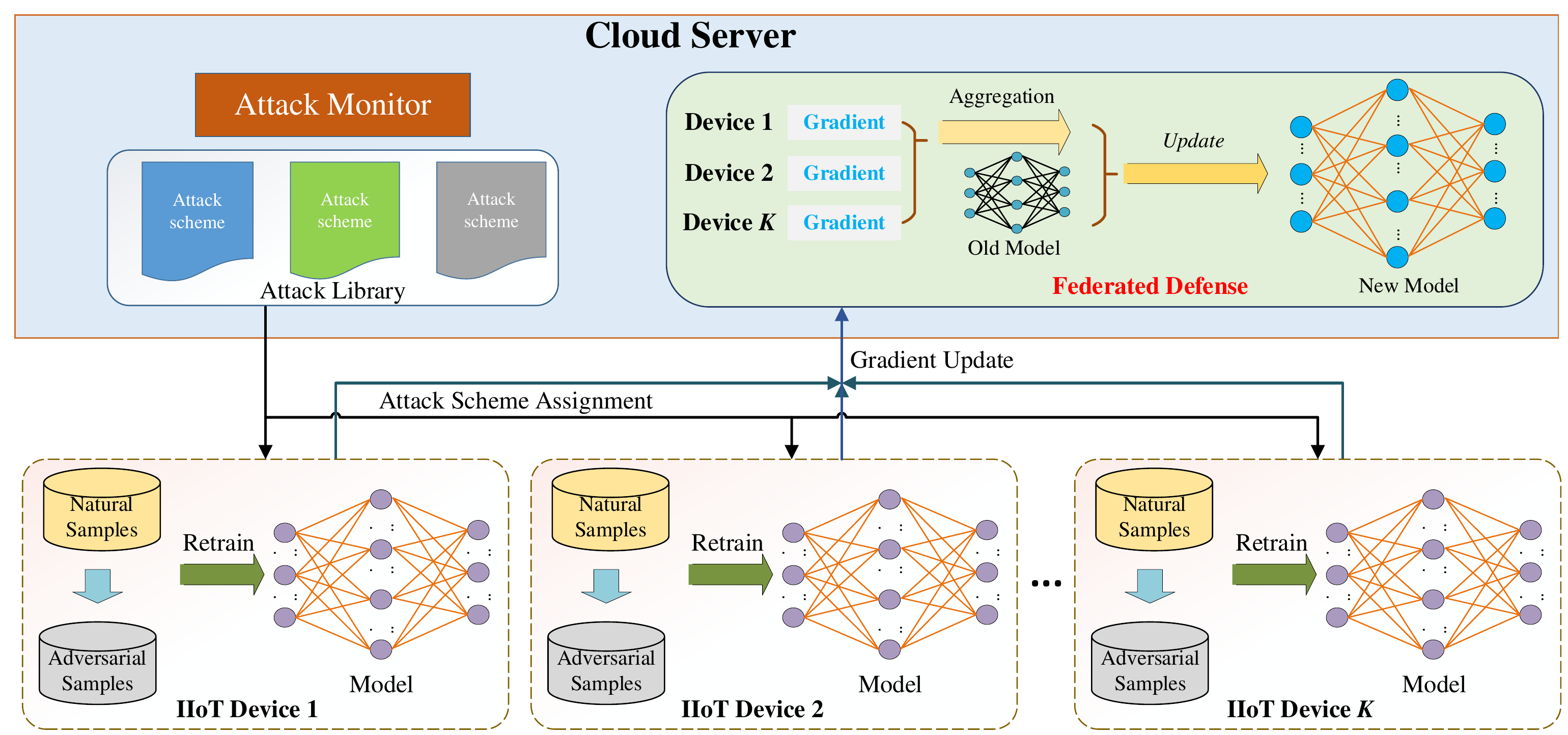}
	\caption{Federated attack detection and defense in FL-based IIoT networks.}
	\label{Fig:Attack}
	\vspace{-0.1in}
\end{figure*}
\textcolor{black}{Industrial devices have become targets of malicious adversaries who can attack AI/ML models in smart manufacturing and operations, by modifying data inputs or changing learning network weights which can lead to erroneous predicted outputs. Many solutions have been proposed to cope with attacks on IIoT devices such as ensemble diversity or adversarial training, but they are mostly applied to a specific type of attack and do not scale well to distributed IIoT networks. FL has emerged as a strong alternative to provide collaborative intelligence for IIoT systems with the ability to detect and prevent various attacks for safe industrial processes.} Enabled by the privacy-promoting feature of FL, a federated attack detection and defense solution is built in \cite{8} where each IIoT machine joins to run a deep neural network locally, in order to retrain the threat model to fight against adversaries. \textcolor{black}{An example for federated attack detection in IIoT is illustrated in Fig.~\ref{Fig:Attack}. Here, each IIoT device first produces adversarial samples to create a retraining set that is then used to build a local attack detection model. Subsequently, the trained gradient is transmitted to the cloud server for aggregation and synchronization to produce a shared model, and this process is iterated in several communication rounds until the attack model converges. In this regard, the built model can effectively detect  attacks thereby building a strong defense solution in the IIoT network.} \textcolor{black}{Through simulations with MINIST datasets, the FL-based approach can achieve a high attack detection accuracy of 87.8\%, compared to 73.8\% in the standalone method.}
\subsection{FL for IIoT Mobile Crowdsensing}
\textcolor{black}{With the rapid development of IIoT, mobile crowdsensing is designed to take advantage of pervasive industrial devices for sensing and collecting data from physical environments. For example, operators in smart factories can make decisions based on environmental information collected from ambient sensors distributed across the whole factory, e.g., sensing abnormal machinery noise to monitor operating status of machines. To realize intelligent mobile crowdsensing, centralized AI/ML techniques are used which usually require direct access to user data, which in turn makes the data vulnerable to privacy leakage. Moreover, the use of a central server to handle all sensing industrial data is not a scalable solution, making it hard to cope with massive data volumes in large-scale industrial systems. FL is a promising tool to accelerate the learning and training for crowdsensing models. As an example, the study in \cite{9} shows an FL-based mobile crowdsensing scheme, with a focus on privacy-enhancing extreme gradient boosting with the cooperation of multiple clients like industrial machines.} A secure gradient aggregation algorithm is designed by integrating homomorphic encryption with secret sharing, which prevents the central server from guessing decryption result before operating aggregation. Simulations  reveal a high accuracy rate of 98\%, and a reduction of 23.9\% runtime, and 33.3\% communication latency for gradient aggregation. 
\begin{figure*}
	\centering
	\includegraphics[width=0.9\linewidth]{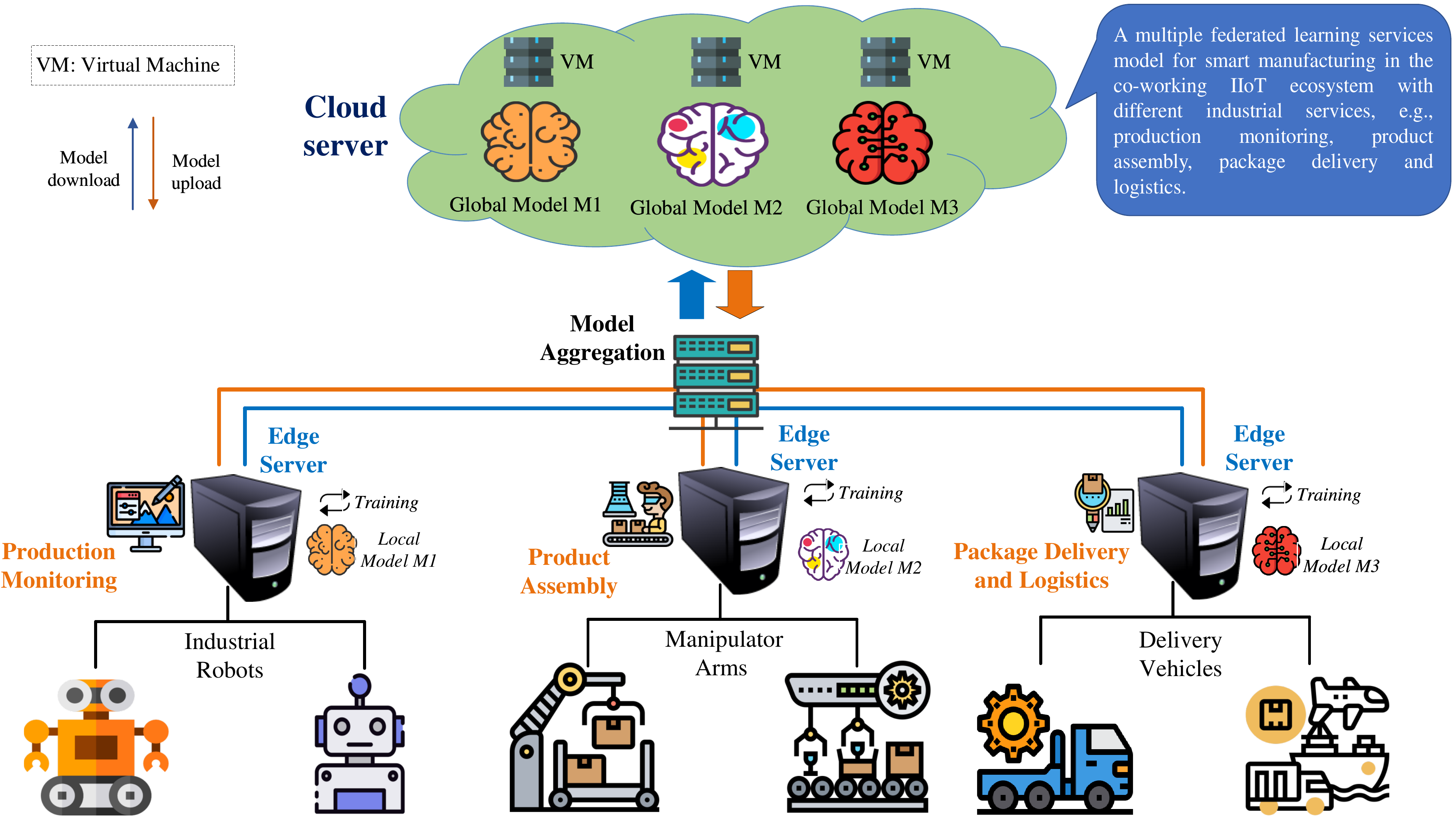}
	\caption{\textcolor{black}{Federated learning for smart manufacturing. }}
	\label{Fig:manufacturing}
	\vspace{-0.1in}
\end{figure*}

\section{FL for IIoT Applications}
\label{Section:App}
In this section, we present the use of FL in IIoT applications in details.  
\subsection{FL for Smart Manufacturing}
Smart manufacturing refers to the integration of intelligence into manufacturing processes where AI techniques play important roles in learning big data generated from industrial machines for process modeling, monitoring, prediction and control in production stages. \textcolor{black}{The AI functions often require data sharing among manufacturers and factories which is not an ideal solution due to growing user privacy concerns.} FL can realize intelligence for industrial systems without data exchange, by the collaborative data learning of distributed industrial devices and machines. 
\textcolor{black}{Given the fact that there are diverse industrial services in industries, e.g., production monitoring with robots, product assembly with automatic manipulator arms, package delivery and logistics with vehicles, it is desired to develop a multiple FL services solution to deal with different industrial services in the co-working IIoT ecosystem, as illustrated in Fig.~\ref{Fig:manufacturing}. Each group of industrial machines participates in the collaborative AI training using their local industrial datasets within their working environments, e.g., machinery fault data in production lines and productivity information in the assembly process, before offloading the learned parameter to the cloud via its edge server. Here, each virtual machine in the cloud will compute a global model of the industrial service that it manages. In this regard, a multiple FL services solution is realized, and all industrial machines in different service lines can benefit from the exchanged knowledge. }

\subsection{FL for Smart Transportation}
Recent advances in sensing and communication technologies along with the growth of data volume from road cameras, embedded devices, and vehicular sensors have empowered vehicular networks. AI/ML has been adopted to realize intelligent transportation systems (ITS) where massive vehicular data are often processed at a data center before sending back to vehicles and roadside units. However, this approach remains some critical issues such as privacy leakage and communication overhead caused by raw data sharing. FL can support ITS by running ML models directly at vehicles based on their datasets such as road geometry, collision avoidance, and traffic flow \cite{5}. A cloud server can be employed to aggregate the local updates of all vehicles to make overall decisions on the traffic flow. The use of massive data from multiple vehicles and huge computation capability of all participants helps provide better traffic prediction outcomes, which cannot be met by using centralized ML techniques with less dataset and limited computation. FL can also support privacy-enhanced smart transport logistics, e.g., package delivery services. In this context, postal operators and customers can federate to run a shared ML model for delivery time prediction based on their local data sources, e.g., traffic conditions, drivers' behaviors, and weather, for delivery latency optimization and thus facilitating logistic activities.
\textcolor{black}{Moreover, FL can be also used in Unmanned Aerial Vehicles (UAVs)-based vehicular networks where UAVs can be employed as mobile FL clients to join the collaborative model training via aerial links with an ITS entity such as a road side unit. This federated mobile model can enable interesting ITS services such as dynamic traffic prediction and road weather monitoring, in which ground-based communications are unavailable.}
\subsection{FL for Smart Grid}
Smart grid plays an integral part in building smart city architectures which not only provides energy resources to smart city applications such as transportation, manufacturing, but also has impacts on environmental, security, and social aspects in Industry 4.0. FL can enable intelligent solutions for smart grid management and energy transmissions in a decentralized manner while helping promote privacy. FL is used to establish federated predictive power schemes in a network of edge data centers for smart grid \cite{11}. In this case, each edge equipment such as a smart meter cooperatively trains AI models, e.g., neural networks, using its own electrical consumption data while the edge server coordinates local updates to build a global model to estimate future household electrical demands. In this way, user information such as energy preference and home addresses is not revealed to the server which promotes privacy protection. Simulations are conducted with over 800 homes in the United States, showing a reduction in networking load, compared to standalone learning approaches. 
\textcolor{black}{Further, due to the increasing privacy concerns, the shareholders  of distributed electric generators and consumers may not willing to provide information of their electricity loads/consumption, but these datasets would be critical to the safe operation of smart grids. FL can help address this issue by allowing the distributed participants to collaboratively learn the patterns of electricity generation/consumption, without sharing raw data to each other. FL is also an efficient solution to bring together different stakeholders from energy systems (heat, cool, gas, etc.), aiming to achieve privacy-enhanced energy information exchange in the electricity production ecosystem. }
\subsection{FL for Smart Healthcare}
In the past few years, AI/ML technologies have been widely used in the healthcare sector to gain insights into health issues and diseases by learning digital medical information extracted from electronic health records (EHRs) for facilitating diagnosis and severity assessment as well as promoting medical research. One of the challenges in such traditional AI techniques is privacy leakage during data analytics. Indeed, compared to other domains, data in healthcare systems are highly sensitive subject to health regulations. Moreover, collecting a large volume of clinical datasets from isolated medical centers is a critical challenge. FL can provide much more efficient solutions for data learning and potentially reshapes the current intelligent healthcare systems by providing intelligent healthcare services while promoting well user privacy based on the cooperation of multiple entities such as health users and healthcare providers across medical institutions. Indeed, FL can offer flexible and privacy-promoting EHRs management solutions \cite{10}, by facilitating the cooperation of multiple hospital institutions to perform health data analytics without the need for EHRs data sharing. \textcolor{black}{Moreover, FL with its privacy-enhanced nature can promote secure healthcare cooperation for better medical service delivery, by allowing for aggregating the model updates from separate hospital organizations with multiple devices, e.g., magnetic resonance imaging (MRI) scanners, to build stronger AI models for medical tasks, such as medical imaging.} 
\begin{figure*}
	\centering
	\includegraphics [width=0.99\linewidth]{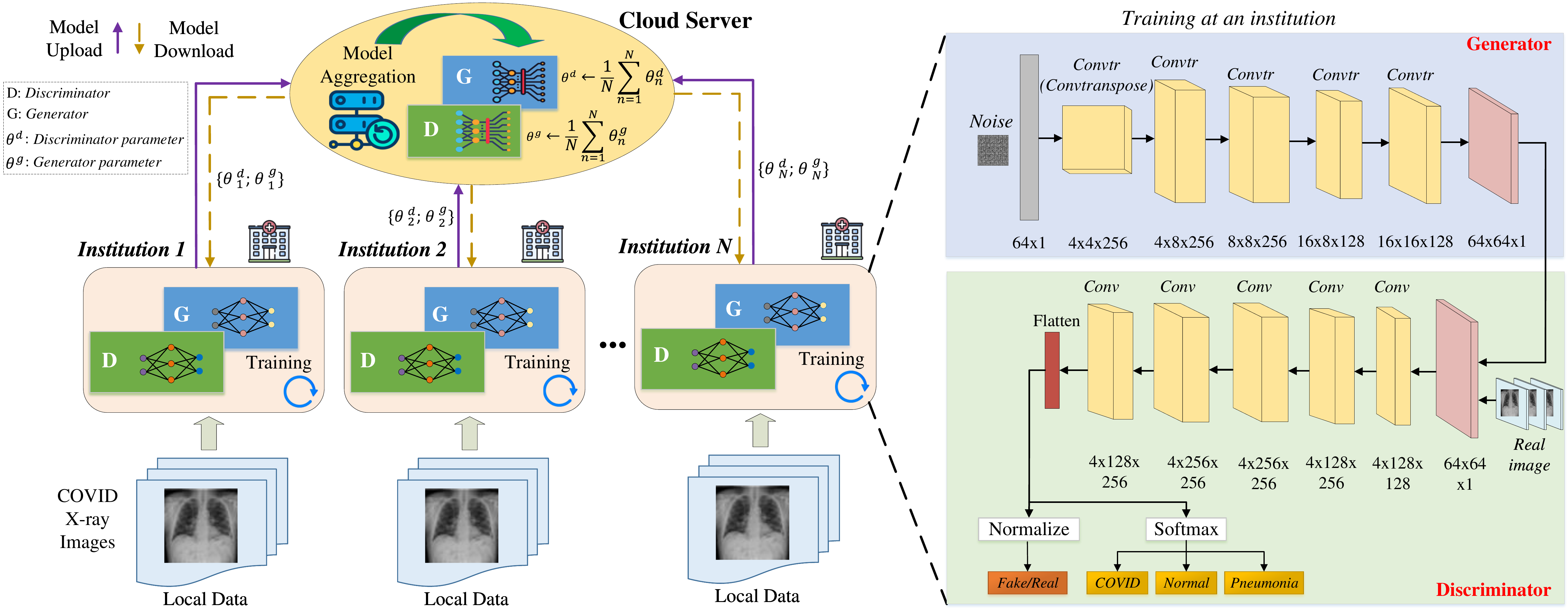}
	\caption{\textcolor{black}{Our advanced FL model for COVID-19 detection.  }}
	\label{Fig:Sharing}
	\vspace{-0.15in}
\end{figure*}

\section{Case Study}
\label{Section:Case}
We present a case study on FL-IIoT, by designing an FL-health system for COVID-19 detection \cite{17}. In the pandemic, collecting sufficient data for training becomes challenging with privacy concerns caused by public data sharing. Hence, we propose a new FL scheme to generate realistic COVID-19 images for facilitating privacy-enhanced COVID-19 detection with generative adversarial networks (GANs) \cite{18}. \textcolor{black}{Compared to the traditional FL scheme \cite{17}, our advanced FL solution can achieve federated data augmentation for generating high-quality synthetic COVID-19 images that can enhance the training performances with privacy awareness.}  The details of our FL design will be provided in the following.
\subsection{System Model}
We consider a system model for FL-based COVID-19 detection as illustrated in Fig.~\ref{Fig:Sharing}, including a set of medical institutions and a cloud server. Each institution participates in the FL process using its own COVID-19 image dataset, \textcolor{black}{e.g., X-ray images,} to build a global GAN with the cloud, aiming to generate high-quality synthetic COVID-19 images for improving the overall COVID-19 detection. \textcolor{black}{Specifically, at each institution we design a GAN consisting of two components, namely a generator and a discriminator based on CNNs which alternatively train via a min-max game \cite{18}. Given a noise sample from a standard Gaussian distribution, the CNN-based generator learns to generate a fake COVID-19 image data point. Moreover, we design another CNN as a discriminator at each institution which tries to classify the real COVID-19 image data point against the one produced from the generator. The discriminator outputs 1 if the input is real data samples or 0 if the input is fake data samples. Accordingly, the generator and the discriminator at each institution interact to obtain the optimal parameters in a fashion that the generator can generate the fake COVID-19 image data distribution close to the real image data as much as possible to fool the discriminator while the discriminator tries to differentiate between fake and real image samples. As a result, the generator can synthesize realistic COVID-19 image samples which are similar to the real COVID-19 image data after a training period, aiming to achieve an efficient data augmentation for later classification tasks.}
\subsection{FL Training for COVID-19 Detection}
\begin{figure*}[t!]
	\centering
	\begin{subfigure}[t]{0.33\textwidth}
		\centering
		\includegraphics[width=0.99\linewidth]{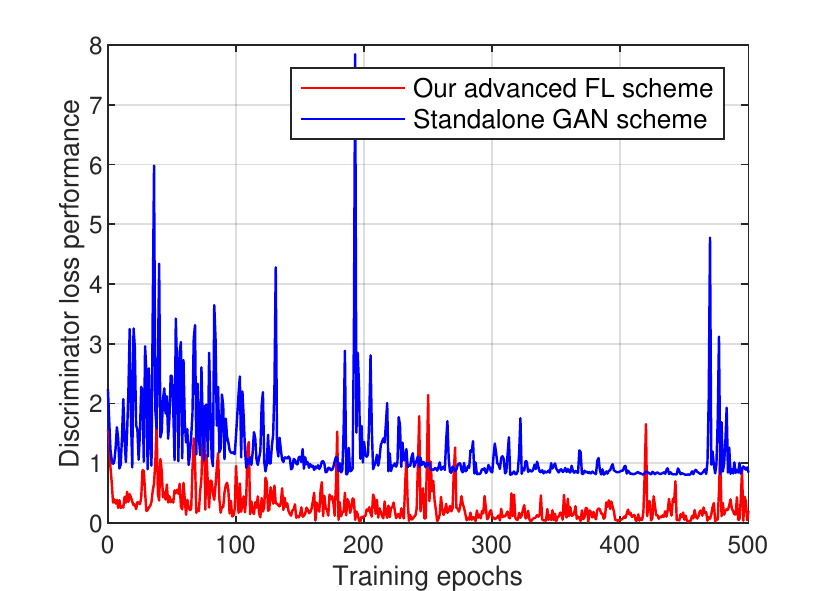} 
		\caption{Performance of discriminator loss.}
	\end{subfigure}%
	~
	\begin{subfigure}[t]{0.33\textwidth}
		\centering
		\includegraphics[width=0.99\linewidth]{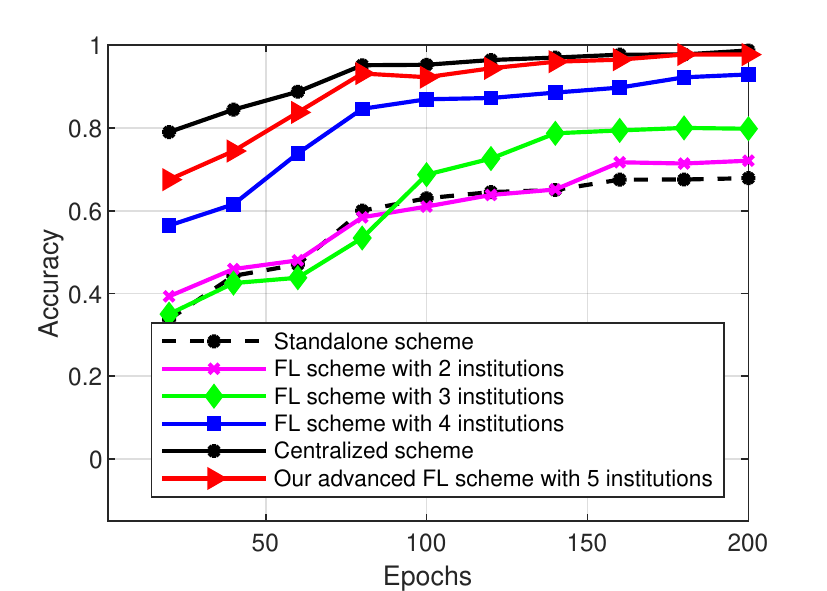} 
		\caption{Performance of COVID-19 detection accuracy. }
	\end{subfigure}%
	~
	\begin{subfigure}[t]{0.33\textwidth}
		\centering
		\includegraphics[width=0.99\linewidth]{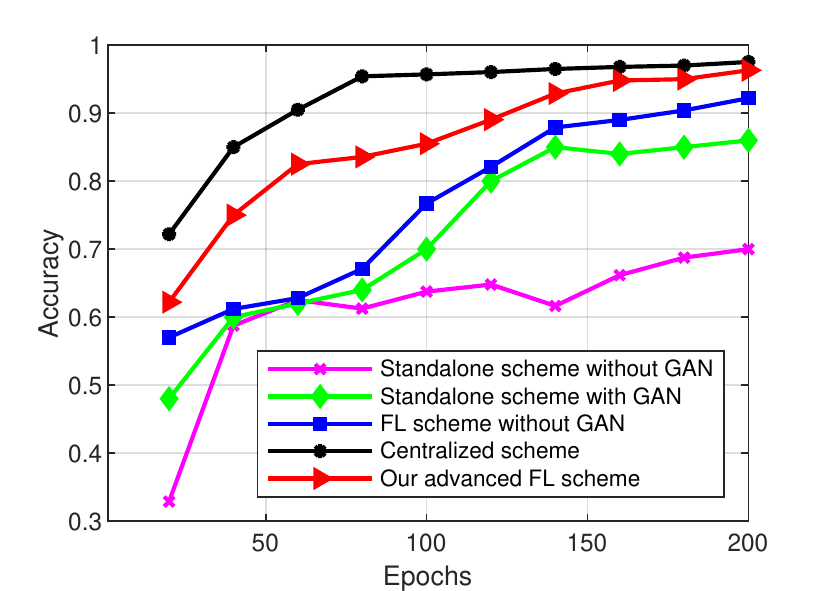} 
		\caption{Performance of COVID-19 detection accuracy. }
	\end{subfigure}
	\caption{Performance comparison of different approaches for COVID-19 detection.}
	\label{Fig:performance}
	\vspace{-0.1in}
\end{figure*}
Each institution joins the FL training with the cloud server, by updating the parameters of the discriminator and the generator in each global round and exchange them with the cloud server for aggregation. For every global epoch, each institution collaboratively trains its discriminator and generator. Specifically, the generator produces minibatchs of fake samples from the noise probability distribution. Also, the discriminator samples minibatchs of real data from the actual image distribution. Then, each institution updates simultaneously the discriminator and generator by ascending its stochastic gradients to update its own weights. After local training, all institutions transmit the learned updates to the cloud server for model averaging, while actual COVID-19 images are kept at local institutions which thus ensures data privacy. Then, the cloud server broadcasts the new global updates to all institutions for the next round of GAN learning. The FL process is iterated until the global loss function converges with a desired accuracy.

\subsection{Illustrative Results} 
We report simulation results obtained when training a COVID-19 dataset \cite{17} of total 620 X-ray images in three classes: COVID-19, normal, and pneumonia in an FL system with five institutions. By using the proposed FL model, we generate 1500 synthetic X-ray images which are then combined with an actual dataset for COVID-19 classification. We use a CNN-based classifier with three convolutional layers and Adam optimizer, and the configurations of GANs are shown in Fig.~\ref{Fig:Sharing}. We evaluate our approach and compare with state-of-the-art schemes, including the standalone scheme (training dataset at only an institution without federation), the standalone scheme with GAN \cite{18}, the FL scheme without GAN \cite{17}, and the centralized scheme.

In Fig.~\ref{Fig:performance}(a), \textcolor{black}{we compare the discriminator loss of our advanced FL scheme and the standalone scheme with GAN \cite{18}}. It can be seen that the the performance of the standalone scheme cannot achieve its optimum due to the lack of access to the full dataset. \textcolor{black}{Meanwhile, our advanced FL scheme} can learn over the entire data span from distributed datasets which is able to extract better image features for efficient data augmentation.

We then investigate the detection accuracy for different FL schemes and our advanced FL scheme, where the standalone scheme is used as the baseline. As shown in Fig.~\ref{Fig:performance}(b), the more participating institutions in data training, the higher accuracy achieved. The intuition behind this observation is the improved image feature learning efficiency thanks to the use of diverse data sources. \textcolor{black}{Nevertheless, the accurate rate of our FL scheme is the best among all approaches and is close to the centralized scheme.}

Additionally, we compare the accuracy performance of our scheme with other COVID-19 detection schemes, as indicated in Fig.~\ref{Fig:performance}(c). Our advanced FL scheme can significantly improve the accuracy performance due to its GAN and federated learning combination. Our scheme yields the highest accuracy of 0.963 after 200 iterative epochs, while other schemes including the FL scheme without GAN, the standalone scheme with GAN, and the standalone scheme without GAN have lower performances, with 0.922, 0.856, and 0.705, respectively. 
\section{Conclusions and Open Research Topics}
\label{Section:Conclude}
\textcolor{black}{This paper provided a detailed overview on the integration of FL into IIoT in industries.} The roles of FL in important IIoT services and applications were identified and analyzed. The feasibility of FL in IIoT was demonstrated via a case study and simulations. Several interesting open research topics for FL-IIoT in industries are highlighted as follows:
\begin{itemize}
	\item \textit{Communication Issues in FL-IIoT:} \textcolor{black}{Communications in FL-IIoT training in both uplinks and downlinks rely heavily on the level of interconnection among machines, AI software, and the computation server. This communication network also differs from traditional ones due to environmental constraints, such as high temperature and corrosive substances in manufacturing processes. \textcolor{black}{Further, the high frequency bands, e.g., above 2.4GHz for WiFi networks, which are essential for low-latency FL communications, may be not available in realistic industrial environments like hospitals}. New designs of efficient communication protocols specific to IIoT settings are desired to facilitate the FL training.}
	
	\item	\textit{Resource Management Issues in FL-IIoT:} The concept of FL-IIoT mostly relies on scalable data parallelism and on-device training at IIoT devices. To achieve a synchronous update at the server, all IIoT devices need to devote their computational resources for training. Unfortunately, this requirement is not always met due to the resource constraints of certain IIoT devices with weak computation capacities, e.g., industrial wearable sensors, which can cause significant delays in model aggregation. Thus, resource-aware FL algorithms and resource allocation solutions should be considered in FL-IIoT system design.
	
	\textcolor{black}{\item	\textit{Economic Issues in FL-IIoT:} In FL-IIoT, when an industrial user serves as training nodes, how to encourage them to join the FL process is a key challenge. A user may not be willing to devote its resources to perform data training if it does not have much economic benefits to compensate the consumption of computational resources. 
	Incentive mechanisms such as credit-based support and revenue payment are highly needed to attract more users to join FL training which also enhances the robustness of industrial FL-IIoT systems.}
\end{itemize}

\section*{Acknowledgments} 
This work was supported in part by the CSIRO Data61, Australia, and in part by U.S. National Science Foundation under Grant CCF-1908308. The work of Jun Li was supported by National Natural Science Foundation of China under Grant 61872184.

\balance
\bibliography{Ref}
\vskip -2\baselineskip 
\begin{IEEEbiographynophoto}{Dinh C. Nguyen}
	\balance	is currently pursuing the Ph.D. degree at the School of Engineering, Deakin University, Victoria, Australia. He is also affiliated with the Information Security and Privacy Research Group, CSIRO Data61, Docklands, Melbourne, Australia.  His research interests focus on federated learning, blockchain, Internet of Things, and edge computing.
\end{IEEEbiographynophoto}

\vskip -2\baselineskip 
\begin{IEEEbiographynophoto}{Ming Ding}
	is currently a Senior Research Scientist with the CSIRO Data61, Sydney, NSW, Australia. His research interests include information technology, data privacy and security, machine learning and AI. He has authored over 100 articles in IEEE journals and conferences. He is an Editor of the IEEE Transactions on Wireless Communications and the IEEE Wireless Communications Letters. 
\end{IEEEbiographynophoto}
\vskip -2\baselineskip 

\begin{IEEEbiographynophoto}{Pubudu N. Pathirana}
	is a full Professor and the Director of Networked Sensing and Control group at the School of Engineering, Deakin University, Geelong, Australia. He was a visiting professor at Yale University in 2009. His current research interests include bio-medical assistive device design, mobile/wireless networks, and Internet of Things. 
\end{IEEEbiographynophoto}

\vskip -2\baselineskip 
\begin{IEEEbiographynophoto}{Aruna Seneviratne}
	is currently a Foundation Professor of telecommunications with the University of New South Wales, Australia, where he holds the Mahanakorn Chair of telecommunications. His current research interests are in physical analytics: technologies that enable applications to interact intelligently and securely with their environment in real time. 
\end{IEEEbiographynophoto}
\vskip -2\baselineskip 
\begin{IEEEbiographynophoto}{Jun Li}
	received Ph. D degree in Electronic Engineering from Shanghai Jiao Tong University, China in 2009. His research interests include network information theory, game theory, distributed intelligence, multiple agent reinforcement learning. He has co-authored more than 200 papers in IEEE journals and conferences, and holds 1 US patents and more than 10 Chinese patents in these areas. He was serving as an editor of IEEE Communication Letters and TPC member for several flagship IEEE conferences.
\end{IEEEbiographynophoto}
\vskip -2\baselineskip 
\begin{IEEEbiographynophoto}{Dusit Niyato}
	(F'17) received the B.Eng. degree from the King Mongkuts Institute of Technology Ladkrabang, Thailand, in 1999, and the Ph.D. degree from the University of Manitoba, Canada, in 2008. He is currently a Professor with the School of Computer Science and Engineering, Nanyang Technological University, Singapore. His research interests are in the areas of energy harvesting for wireless communication, the Internet of Things, and sensor networks.
\end{IEEEbiographynophoto}
\vskip -2\baselineskip 
\begin{IEEEbiographynophoto}{H. Vincent Poor}
	(F'87) is the Michael Henry Strater University Professor of Electrical Engineering at Princeton University. His interests include information theory, machine learning and networks science, and their applications in wireless networks, energy systems, and related fields. Dr. Poor is a Member of the National Academy of Engineering and the National Academy of Sciences, and a Foreign Member of the Chinese Academy of Sciences and the Royal Society. He received the Marconi and Armstrong Awards of the IEEE Communications Society in 2007 and 2009, respectively, and the IEEE Alexander Graham Bell Medal in 2017.
\end{IEEEbiographynophoto}

\bibliographystyle{IEEEtran}
\end{document}